\documentclass[11pt]{article}

\usepackage[final]{acl}

\usepackage{times}
\usepackage{latexsym}

\usepackage[T1]{fontenc}

\usepackage[utf8]{inputenc}

\usepackage{microtype}

\usepackage{inconsolata}

\usepackage{graphicx}
\usepackage{subcaption}

\usepackage{booktabs}

\usepackage{fancyvrb}

\usepackage[most]{tcolorbox}
\definecolor{takeawaybg}{HTML}{EEF3FA}
\definecolor{takeawayrule}{HTML}{4A6FA5}
\newtcolorbox{takeaway}[1]{enhanced, breakable, colback=takeawaybg, colframe=takeawaybg, boxrule=0pt,
  borderline west={2pt}{0pt}{takeawayrule}, sharp corners, left=5pt, right=4pt, top=2pt, bottom=2pt,
  before skip=4pt, after skip=6pt, before upper={\textbf{#1}\par}}
\newcommand{\ph}[1]{\textrm{\ensuremath{\langle}\textit{#1}\ensuremath{\rangle}}}

\usepackage{amsmath}
\usepackage{amssymb}
\usepackage{amsthm}
\newcommand{\method}{Runoff}

\newtheorem{proposition}{Proposition}

\theoremstyle{definition}
\newtheorem{definition}{Definition}

\title{Consider, Then Choose: Order-Invariant Decisions in System One Models}

\author{
  Jianguo Huang\thanks{Equal contribution.} \quad
  Yao Long Teng\footnotemark[1] \quad
  Yanchen Deng\thanks{Corresponding author.} \quad
  Bo An \\
  Nanyang Technological University, Singapore
}

\begin{document}
\maketitle

\begin{abstract}
System One models answer typed questions in a single forward pass and promise the same answer for the same input. For choice questions, this promise fails: the answer depends on the order of the options, increasingly so as they grow in number, and averaging over their cyclic shifts reduces this dependence without removing it. Scoring each option alone, which the order cannot affect, shows that the order changes the answer mostly when the model is unsure among its top few options, even though the listwise answer looks confident. We therefore propose \textbf{Runoff}, which considers, then chooses: it scores each option alone to keep a few candidates, then compares them in listings where each candidate takes every position. It needs no training and can run in two forward passes, and its answer does not depend on the option order by construction. On two open models and five datasets, it changes no answer under shuffling, against 9.3\% for cyclic averaging, and with 20 options it is about as accurate and three to four times faster. Wrapped around eight System One models, it removes the order dependence of the five benchmarks, while cyclic averaging changes up to 15.4\%.
\end{abstract}

\section{Introduction}
\label{sec:intro}

System One decision models answer typed questions without generating
text~\citep{typesafe2026jev}: given a state and a question whose answer is a yes/no, a choice among several options, or a level on an ordered scale, they return a probability distribution over the answers.
A language model can be asked the same questions in free text, but the
caller must then parse the response, and any probabilities it exposes
are over tokens rather than over the possible answers.
In real applications, an agent system has to make fast decisions, such as routing a request among tools or picking the next element on a screen.
Fast decision models have thus become popular, and open implementations now include both purpose-trained decision models and readout heads on frozen LLMs~\citep{llm2jev2026,semif2026,laya2026,anyjev2026}.

\begin{figure*}[!t]
\centering
\begin{subfigure}[t]{0.24\textwidth}
\centering
\includegraphics[width=\linewidth]{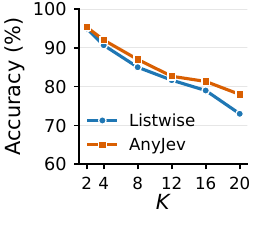}
\caption{Banking20, accuracy}
\end{subfigure}\hfill
\begin{subfigure}[t]{0.24\textwidth}
\centering
\includegraphics[width=\linewidth]{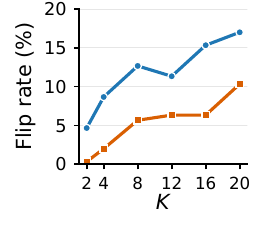}
\caption{Banking20, flip rate}
\end{subfigure}\hfill
\begin{subfigure}[t]{0.24\textwidth}
\centering
\includegraphics[width=\linewidth]{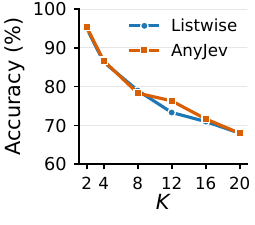}
\caption{Newsgroups, accuracy}
\end{subfigure}\hfill
\begin{subfigure}[t]{0.24\textwidth}
\centering
\includegraphics[width=\linewidth]{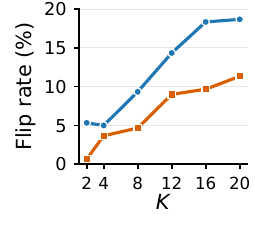}
\caption{Newsgroups, flip rate}
\end{subfigure}
\caption{Accuracy and flip rate against the number of options, Qwen3-8B. Flip rate: share of answers that change when the options are shuffled. }
\label{fig:k}
\end{figure*}

For choice questions, however, the answer depends not only on the question’s content but also on the order in which the options are listed.
Jev's own documentation acknowledges that models tend to favor the first-listed option, and advises users to reorder the options and check that the answer stays the same~\citep{typesafe2026jaggedness}.
Moreover, the effect grows with the number of options. As shown in Figure~\ref{fig:k}, when Qwen3-8B answers a 20-way intent question by scoring the option letters, shuffling the options changes 17\% of its answers, compared with 5\% for two-way questions.
This matters because in an agent the option order is seldom a deliberate choice, but is inherited from the order of a tool registry or the layout of a page.
An answer that depends on this order can change whenever one of these components changes, and anyone who controls the order can steer the answer~\citep{pezeshkpour-hruschka-2024-large}.

Existing methods do not remove the dependence.
Asking the model to reason in text before answering~\citep{wang-etal-2024-large-language-models-fair} gives up the single forward pass that defines a System One model, and on choice questions it does not reliably make the answer independent of the order \citep{pezeshkpour-hruschka-2024-large}.
Averaging over the $K$ cyclic shifts of the options~\citep{zheng2024large,anyjev2026} reduces the dependence but does not remove it: on open models, it still changes the answers under a shuffle.
It is also costly, with cost growing quadratically with $K$.

In Section~\ref{sec:motivation}, we locate this dependence by comparing the listwise answer with pointwise decisions of the same model, which score each option alone and are therefore blind to the order.
This comparison shows where and when the order matters.
The flips occur among a few options: at least one of the answers before and after a flip almost always lies among the options that the pointwise decisions rank highest.
The order matters most when the model is uncertain: a shuffle changes far more answers when the pointwise top option is uncertain than when it is certain, even though the listwise answer looks confident in both cases.
As more options make the model uncertain more often, the dependence grows with the number of options.

These findings suggest the two steps that people take when facing many options: consider, then choose.
Shoppers narrow the available brands to a consideration set before choosing among them \citep{hauser1990evaluation}, much as an election holds a runoff among the leading candidates of a first round.
Our method, \method{}, brings these two steps to System One models, using only the decisions of the model itself.
It first keeps the options that the pointwise decisions rank highest, so the ruled-out options cannot interfere with the choice.
It then compares these candidates in every cyclic shift of their pointwise ranking, so each candidate takes every position once and none is favored by its place.
Since neither step uses the input order, the answer of \method{} does not depend on it by construction, and the method needs no training and can run in two forward passes.

In Section~\ref{sec:experiments}, we evaluate \method{} on two open models, Qwen3-8B and Gemma 4 12B, over five datasets with 3 to 20 options, and as a black box around eight System One models.
On the open models, it changes no answers under shuffling, whereas cyclic averaging changes 9.3\% and a single listwise decision in the given order (Listwise) 18.6\%.
With 20 options, it is about as accurate as cyclic averaging and several times faster.
Its main limitation is reasoning questions, where the consideration set more often misses the correct answer and it trails cyclic averaging by a few points.
Around the System One models, all of which change their own answers under shuffling, it removes the order dependence of every model that returns the same probabilities for a repeated request, and leaves only a small residue on the others, far below that of cyclic averaging, with similar accuracy.

Our contributions are:
\begin{itemize}
\item An analysis showing that option-order dependence grows with the number of options, and that the order matters most when the model is unsure among its top few options.
\item \method{}, a training-free decision rule that first narrows the options and then compares only the leading few, which makes its answer independent of the option order and cheaper than cyclic averaging when the options are many.
\item Experiments on ten models and five datasets, where \method{} removes the order dependence of every deterministic model with accuracy close to cyclic averaging, and falls behind it only on reasoning questions.
\end{itemize}

\section{Preliminaries}
\label{sec:prelim}

\paragraph{Notation.}
We consider a System One decision model \citep{almeida2026jev} answering a choice question $q$ about a state $x$, with a set $\mathcal{O}$ of $K$ options, exactly one of which is correct.
A listing of $\mathcal{S} \subseteq \mathcal{O}$ is a sequence $\lambda = (\lambda(1), \dots, \lambda(|\mathcal{S}|))$ that contains every option of $\mathcal{S}$ once, which shows $\lambda(i)$ at position $i$.
The question presents its options in its own \emph{order} $\sigma$, a listing of $\mathcal{O}$. A method may make its decisions on $\sigma$ or on any listing of $\mathcal{O}$.

\paragraph{Listwise decision.}
The listwise decision on a listing $\lambda$ shows its options under the letters A, B, \dots{} and renormalizes the next-token probabilities of these letters~\citep{robinson2023leveraging}.
Made on the order of the question, $P_\sigma = g(x, q, \sigma)$ is the standard answer, which we call \emph{Listwise}.

\paragraph{Cyclic averaging.}
AnyJev \citep{anyjev2026} makes the listwise decision on the $K$ rotations of the order, $\sigma_s = (\sigma(s), \dots, \sigma(K), \sigma(1), \dots, \sigma(s-1))$, which place every option at every position once, and averages their log-probabilities:
\begin{equation}
P^{\textsc{cyc}}_\sigma(o) \propto \exp\Big(\frac{1}{K}\sum_{s=1}^{K} \log g(x, q, \sigma_s)(o)\Big).
\label{eq:cyc}
\end{equation}

Next, we give the definition of order invariance, which is the property that we want to achieve.
\begin{definition}[Order invariance]
\label{def:invariance}
A decision rule $R$, which maps $(x, q, \sigma)$ to a distribution $R(x, q, \sigma)$ over $\mathcal{O}$, is order-invariant if $R(x, q, \sigma) = R(x, q, \sigma')$ for all $x$, $q$, and orders $\sigma, \sigma'$.
\end{definition}
Listwise is not order-invariant, since $\sigma$ decides which letter each option receives.
Neither is AnyJev: it places every option at every position, but its $K$ listings are the rotations of $\sigma$, so $\sigma$ decides which $K$ of the $K!$ listings are averaged.
More details of these methods are shown in Appendix~\ref{app:prompt}.

\begin{figure*}[t]
\centering
\begin{subfigure}[t]{0.32\textwidth}
\centering
\includegraphics[width=\linewidth]{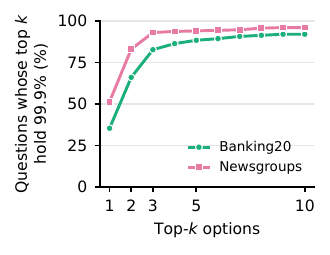}
\caption{Long tail of the pointwise distribution}
\label{fig:head-tail}
\end{subfigure}\hfill
\begin{subfigure}[t]{0.32\textwidth}
\centering
\includegraphics[width=\linewidth]{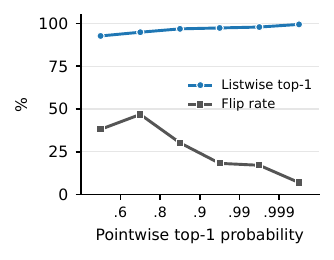}
\caption{Confidence and flips of Listwise}
\label{fig:head-conf}
\end{subfigure}\hfill
\begin{subfigure}[t]{0.32\textwidth}
\centering
\includegraphics[width=\linewidth]{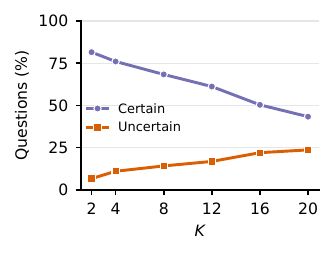}
\caption{Certainty against $K$}
\label{fig:head-sure}
\end{subfigure}
\caption{Order dependence of Listwise and its relation to the pointwise distribution for Qwen3-8B. Panels (b) and (c) pool Banking20 and Newsgroups. (a) Percentage of questions whose top k pointwise options hold 99.9\% of the probability, at K=20. (b) Mean top-1 probability of Listwise and its flip rate at K=20, with questions binned by their pointwise top-1 probability, which does not depend on the order. (c) Percentage of uncertain (pointwise top-1 probability below 0.9) and certain (at least 0.999) questions as K grows.}
\label{fig:head}
\end{figure*}

\section{Motivation and Analysis}
\label{sec:motivation}

This section studies how the option order affects the answers of Jev-style decision models.
Following AnyJev~\citep{anyjev2026}, we turn an open-weight LLM into such a model: it reads the state and the question in a single forward pass, without decoding, and returns the renormalized next-token probabilities of the option letters as its answer distribution.
We answer each question with Listwise and AnyJev, once with the options in their original order and once in a shuffled order, and measure the order dependence by the \emph{flip rate}, the percentage of questions whose answer changes between the two orders.

\textbf{Flip rate and the number of options.} To study the effect of the number of options, we vary $K$ from 2 to 20 and measure the flip rate and accuracy of Listwise and AnyJev.
We take two 20-way classification tasks, a 20-intent subset of BANKING77~\citep{casanueva-etal-2020-efficient} and 20 Newsgroups~\citep{lang1995newsweeder}: the gold label and the first $K-1$ of the other labels in a fixed random order.
Figure~\ref{fig:k} shows the results for Qwen3-8B \citep{yang2025qwen3}.
With Listwise, the flip rate rises from 4.7\% at $K{=}2$ to 17.0\% at $K{=}20$ on Banking20 and from 5.3\% to 18.7\% on Newsgroups, while accuracy falls.
AnyJev lowers the flip rate but does not stop its growth, from 0.3\% to 10.3\% and from 0.7\% to 11.3\%.

These results show that changing the order of options makes the predictions unreliable.
This raises the question: \textit{Where and when does the order change the answer?}
To find out, we need the model's belief about each option without any order, which we measure with pointwise decisions.

\textbf{Pointwise decision.}
The pointwise decision compares the model's belief about each option alone, as shown in Appendix~\ref{app:prompt}:
\begin{equation}
f(x, q, o) = \log P(\text{yes} \mid x, q, o) - \log P(\text{no} \mid x, q, o).
\end{equation}
Then, the \emph{pointwise distribution} is given by normalizing the odds:
\begin{equation}
\pi(o) = \frac{\exp f(x, q, o)}{\sum_{o' \in \mathcal{O}} \exp f(x, q, o')}, \qquad o \in \mathcal{O}.
\label{eq:pi}
\end{equation}
Each decision sees one option alone, so $\pi$ does not depend on $\sigma$: it shows how certain the model is of each option.
The pointwise distribution is highly concentrated on a few options (Figure~\ref{fig:head-tail}): at $K{=}20$, the top three options of Qwen3-8B hold 99.9\% of its probability for 83\% (Banking20) and 93\% (Newsgroups) of the questions.
We call these three options the \emph{candidates}.

\begin{table}[t]
\centering
\small
\begin{tabular}{lccc}
\toprule
 & Both & One & Neither \\
\midrule
\multicolumn{4}{l}{\emph{Qwen3-8B}} \\
Banking20 & 43.1 & 47.1 & 9.8 \\
Newsgroups & 80.4 & 17.9 & 1.8 \\
\midrule
\multicolumn{4}{l}{\emph{Gemma 4 12B}} \\
Banking20 & 69.6 & 30.4 & 0.0 \\
Newsgroups & 72.1 & 20.9 & 7.0 \\
\bottomrule
\end{tabular}
\caption{Share of flips (\%) with both, one, or neither of the two answers among the three candidates, $K{=}20$.}
\label{tab:flip-rank}
\end{table}

\textbf{Where the answer moves.}
For each flip, we check whether the answers under the original and the shuffled order are among the three candidates. As Table \ref{tab:flip-rank} shows, in over 90\% of the flips at least one of the two answers is a candidate, and often both are.
Neither is a candidate in at most 10\% of the flips.
The order thus mostly decides among the few options that the model ranks highest.

\textbf{Flip rate and pointwise certainty.}
To see when the order affects the answers, we group the $K{=}20$ questions by the top-1 probability of $\pi$, which does not depend on the order and measures how certain the model's prediction is (Figure~\ref{fig:head-conf}).
The flip rate grows as the model becomes less certain: it exceeds a third below 0.8, is 30\% between 0.8 and 0.9, about 17\% between 0.9 and 0.999, and 7\% at 0.999 and above.
Meanwhile, the top-1 probability of Listwise barely moves, from 93\% to 99.5\%.

\textbf{Certainty and the number of options.}
To see why the flip rate grows with $K$, we track how certain the model’s prediction is as the number of options increases. Figure~\ref{fig:head-sure} shows that as $K$ grows from 2 to 20, the share of uncertain questions rises from 7\% to 24\%, while the share of \emph{certain} questions, whose pointwise top-1 probability is at least 0.999, falls from 82\% to 43\%.
Since the order changes the answer more often the more uncertain the prediction is, this shift toward uncertainty makes the flip rate grow with the number of options.

\begin{takeaway}{Takeaways}
\begin{list}{\textbullet}{\setlength{\leftmargin}{1.1em}\setlength{\labelwidth}{0.7em}\setlength{\labelsep}{0.4em}%
  \setlength{\topsep}{1pt}\setlength{\partopsep}{0pt}\setlength{\itemsep}{0pt}\setlength{\parsep}{0pt}}
\item \emph{How much:} The answer depends on the option order, more so with more options, and AnyJev does not stop it.
\item \emph{Where:} Most answers before and after a flip are among the model's top three options.
\item \emph{When:} The more uncertain the prediction, the more often the order changes the answer.
\item \emph{Why it grows:} More options make the prediction more uncertain, so the order matters more.
\end{list}
\end{takeaway}

\begin{figure*}[!t]
\centering
\includegraphics[width=\linewidth]{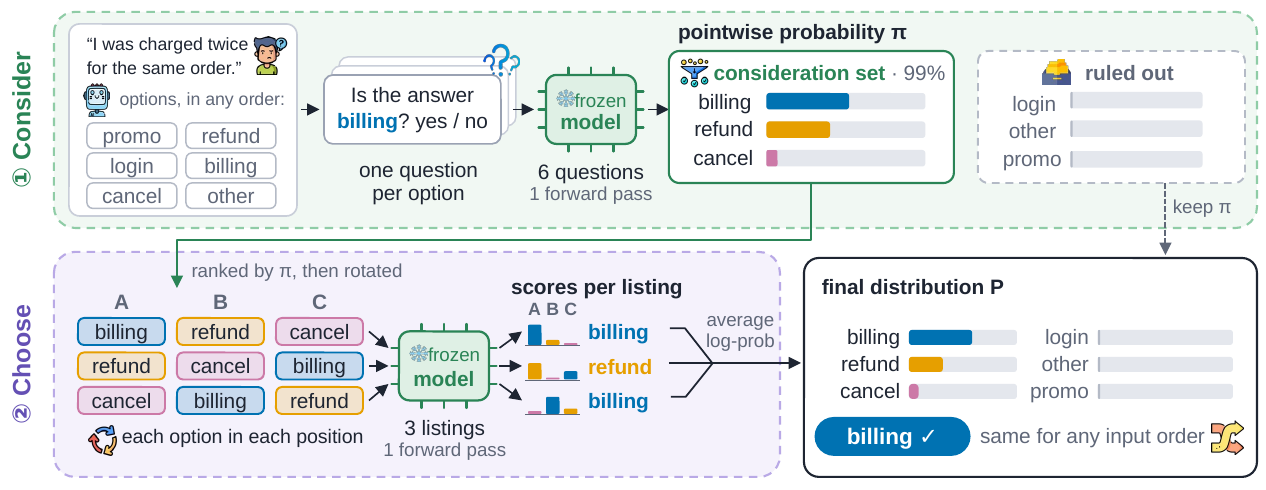}

\caption{Overview of Runoff.}
\label{fig:framework}
\end{figure*}

\section{Method}
\label{sec:method}

We propose \method{}, a decision rule that needs no training and whose answer does not depend on the option order.
As Figure \ref{fig:framework} illustrates, Runoff works in two steps. Consider picks the few options among which the order can move the answer, by scoring each option alone as in Section \ref{sec:motivation}. Choose then compares only these options, placing each at every position.
Unlike cyclic averaging, which compares all $K$ options in $K$ listings, \method{} compares only these few, so the options it rules out cannot change the answer (Appendix~\ref{app:shuffles}).

\textbf{Consider.}
\method{} computes the pointwise distribution $\pi$, defined in Eq.~\ref{eq:pi}, and keeps its top options as a \emph{consideration set} $\mathcal{C}$: the smallest set of top options that holds at least $\alpha$ of the probability and contains at least $k_{\min}$ options.
Formally, let $v_1 > \dots > v_n$ be the distinct values of $\pi$ and $U_j = \{o \in \mathcal{O} : \pi(o) \ge v_j\}$. Then, $\mathcal{C} = U_J$ with $J = \min\{j : \pi(U_j) \ge \alpha,\ |U_j| \ge k_{\min}\}$, and $\mathcal{C} = \mathcal{O}$ if no such $j$ exists. Thus, the consideration set grows when the model’s prediction is uncertain.

\textbf{Choose.}
The pointwise distribution alone is order-invariant, but it judges each option without seeing the others, which leads to wrong answers.
Choose therefore compares the members of $\mathcal{C}$ side by side.
Specifically, we first rank them by decreasing $\pi$. We call the result $c_1, \dots, c_m$ the \emph{canonical ranking}.
A single listwise decision on this ranking would already be order-invariant, but the model's position bias would then favor whichever candidate the ranking puts first, i.e., the top option of $\pi$.
Choose instead makes a listwise decision on each of the $m$ cyclic shifts of the ranking, $\lambda_s = (c_s, \dots, c_m, c_1, \dots, c_{s-1})$, so that every candidate takes every position once: for a ranking $(a, b, c)$, the listings are $(a, b, c)$, $(b, c, a)$ and $(c, a, b)$.
It averages their log-probabilities as in Eq.~\ref{eq:cyc}, which gives a distribution $r$ over $\mathcal{C}$.
The final distribution keeps $\pi$ outside $\mathcal{C}$ and divides the probability of $\mathcal{C}$ according to $r$:
\begin{equation}
P(o) = \begin{cases} \pi(\mathcal{C})\, r(o) & o \in \mathcal{C}, \\ \pi(o) & \text{otherwise}. \end{cases}
\label{eq:runoff}
\end{equation}

\begin{proposition}\label{prop:invariance}
\method{} is order-invariant (Definition~\ref{def:invariance}): the requests it sends to the model, and hence $P$, are the same for every order $\sigma$ of $\mathcal{O}$.
\end{proposition}

\noindent
Each pointwise decision sees one option and $\pi$ normalizes over the set $\mathcal{O}$, so $\pi$ depends only on $(x, q, \mathcal{O})$; $\mathcal{C}$, its canonical ranking and the cyclic shifts are functions of $\pi$ and the option texts (proof in Appendix~\ref{app:proofs}).
The invariance holds whatever position bias the model has, and on a black-box model any remaining flip can only come from the model answering identical requests differently.

\textbf{Cost.}
\method{} reads $K + m^2$ option slots, against $K^2$ for cyclic averaging, so it is cheaper when $m \ll K$.
On open models, each stage runs as a single forward pass.

\begin{table*}[t]
\centering
\footnotesize
\setlength{\tabcolsep}{3.5pt}
\renewcommand{\arraystretch}{0.88}
\begin{tabular}{llcccccccccccc}
\toprule
 & & \multicolumn{2}{c}{Banking20} & \multicolumn{2}{c}{Newsgroups} & \multicolumn{2}{c}{MMLU-Pro} & \multicolumn{2}{c}{Typed Decisions} & \multicolumn{2}{c}{JevBench} & \multicolumn{2}{c}{Avg} \\
\cmidrule(lr){3-4} \cmidrule(lr){5-6} \cmidrule(lr){7-8} \cmidrule(lr){9-10} \cmidrule(lr){11-12} \cmidrule(lr){13-14}
Model & Method & Acc$\uparrow$ & Flip$\downarrow$ & Acc$\uparrow$ & Flip$\downarrow$ & Acc$\uparrow$ & Flip$\downarrow$ & Acc$\uparrow$ & Flip$\downarrow$ & Acc$\uparrow$ & Flip$\downarrow$ & Acc$\uparrow$ & Flip$\downarrow$ \\
\midrule
\multicolumn{14}{l}{\textit{Open-source}} \\
\midrule
Qwen3-8B & Listwise & 73.0 & 17.0 & 68.0 & 18.7 & 47.6 & 44.4 & 62.7 & 12.2 & 66.2 & 21.6 & 63.5 & 22.8 \\
 & AnyJev & \textbf{78.0} & 10.3 & 68.0 & 11.3 & \textbf{53.1} & 24.7 & \textbf{65.2} & 3.0 & \textbf{69.8} & 8.6 & \textbf{66.8} & 11.6 \\
 & \method{} (ours) & 76.7 & \textbf{0.0} & \textbf{68.3} & \textbf{0.0} & 51.6 & \textbf{0.0} & 64.2 & \textbf{0.0} & 68.3 & \textbf{0.0} & 65.8 & \textbf{0.0} \\
\midrule
Gemma 4 12B & Listwise & 83.7 & 7.7 & 67.3 & 14.3 & 57.1 & 31.9 & 70.0 & 11.5 & \textbf{87.1} & 6.5 & 73.0 & 14.4 \\
 & AnyJev & \textbf{84.3} & 3.0 & 67.7 & 7.3 & \textbf{60.1} & 16.2 & \textbf{70.8} & 5.8 & 86.3 & 2.9 & \textbf{73.8} & 7.1 \\
 & \method{} (ours) & 81.3 & \textbf{0.0} & \textbf{68.7} & \textbf{0.0} & 57.0 & \textbf{0.0} & 69.3 & \textbf{0.0} & 84.9 & \textbf{0.0} & 72.2 & \textbf{0.0} \\
\midrule
\multicolumn{14}{l}{\textit{Closed-source}} \\
\midrule
Jev 1.13 & Native & 86.7 & 3.0 & \textbf{75.7} & 5.0 & 81.8 & 7.7 & 72.0 & 6.5 & 87.1 & 4.3 & 80.6 & 5.3 \\
 & AnyJev & \textbf{87.7} & \textbf{1.3} & 75.0 & 1.7 & \textbf{83.6} & 3.4 & 72.0 & 3.7 & \textbf{89.2} & 0.7 & \textbf{81.5} & 2.2 \\
 & \method{} (ours) & 87.0 & \textbf{1.3} & 75.3 & \textbf{0.3} & 83.4 & \textbf{1.4} & \textbf{72.8} & \textbf{1.5} & \textbf{89.2} & \textbf{0.0} & \textbf{81.5} & \textbf{0.9} \\
\midrule
Mercury Decide & Native & 89.7 & 3.7 & \textbf{74.0} & 5.3 & 63.5 & 22.3 & 67.3 & 9.8 & 89.2 & 7.9 & 76.7 & 9.8 \\
 & AnyJev & \textbf{91.0} & 2.3 & 73.7 & 4.3 & \textbf{68.6} & 12.3 & 68.3 & 5.2 & 90.6 & 2.9 & 78.4 & 5.4 \\
 & \method{} (ours) & \textbf{91.0} & \textbf{0.0} & 73.7 & \textbf{0.7} & 66.7 & \textbf{4.4} & \textbf{69.2} & \textbf{0.7} & \textbf{92.1} & \textbf{1.4} & \textbf{78.5} & \textbf{1.4} \\
\midrule
GPT-6 Luna & Native & 85.5 & 5.9 & \textbf{75.2} & 6.7 & 61.8 & 27.7 & 65.5 & 18.7 & 89.2 & 7.2 & 75.4 & 13.2 \\
 & AnyJev & \textbf{87.9} & 2.8 & 74.8 & 3.0 & \textbf{64.8} & 15.0 & 67.0 & 9.8 & \textbf{91.4} & 2.2 & \textbf{77.2} & 6.6 \\
 & \method{} (ours) & 85.9 & \textbf{0.0} & \textbf{75.2} & \textbf{0.0} & 64.6 & \textbf{0.0} & \textbf{68.8} & \textbf{0.0} & 90.6 & \textbf{0.0} & 77.0 & \textbf{0.0} \\
\bottomrule
\end{tabular}
\caption{Main results on five decision benchmarks. Acc: accuracy (\%) with the options in their original order. Flip: share of questions (\%) whose answer changes when the options are shuffled. Native: the model's own choice answer. Avg: mean over the five datasets. Best per model in bold. }
\label{tab:main}
\end{table*}

\section{Experiments}
\label{sec:experiments}

\subsection{Setup}
\label{sec:setup}

\textbf{Decision Models.} We evaluate ten models in three groups: (i) open-weight LLMs read out as decision models, Qwen3-8B \citep{yang2025qwen3} and Gemma 4 12B \citep{gemmateam2026gemma4};
(ii) closed-source System One models, Jev 1.13, Mercury Decide and GPT-6 Luna; (iii) open-weight models fine-tuned for System One decisions, Torchcast 12B, Kev 9B, Intern-Decision 4B, Tev1 4B and Laya.

\textbf{Evaluation.}
We evaluate the choice questions of five benchmarks: Banking20 and Newsgroups (20 options, 300 questions each), MMLU-Pro \citep{wang2024mmlupro} (10 options; 1,400 questions), Typed Decisions \citep{typeddecisions2026} (600 questions with 4--5 options), and the public items of JevBench \citep{jevbench2026} (139 questions with 3--6 options).
The baselines are Listwise and cyclic averaging as implemented by AnyJev. The prompt templates are provided in Appendix~\ref{app:prompt}. 
For \method{}, we use $\alpha = 0.99$ and $k_{\min} = 3$.

\textbf{Evaluation metrics.}
We report accuracy with the options in their original order (Acc) and the flip rate under one shuffle per question (Flip).

\subsection{Main results}
\label{sec:results}

\paragraph{Results of open-source models on five decision benchmarks.}
Table~\ref{tab:main} shows the results of the open-source models, i.e., Qwen3-8B and Gemma 4 12B.
\method{} changes no answers when the options are shuffled, whereas Listwise and AnyJev change theirs on every dataset, and its accuracy stays comparable to that of both baselines.
For example, on MMLU-Pro with Qwen3-8B, shuffling changes 44.4\% of the Listwise answers and 24.7\% of the AnyJev answers, but none of \method{}'s.
In accuracy, with 20 options, \method{} reaches 73.8\% averaged over Banking20 and Newsgroups on both models, against 74.5\% for AnyJev and 73.0\% for Listwise.
Overall, \method{} removes the order dependence while keeping its accuracy comparable to that of both baselines.

\paragraph{Results of closed-source models.}
Table~\ref{tab:main} also reports the results of the three closed-source System One models, with AnyJev and \method{} wrapped around their APIs.
The native answers of all three models depend on the option order, but \method{} reduces this dependence further than AnyJev on each of them, and on average it is at least as accurate.
For example, GPT-6 Luna answers a re-sent request identically, yet shuffling changes 13.2\% of its native answers and 6.6\% of AnyJev's on average, but none of \method{}'s, while the accuracy rises from 75.4\% to 77.0\% (AnyJev: 77.2\%).
On Jev 1.13 and Mercury Decide, \method{} lowers the flip rate from 5.3\% and 9.8\% to 0.9\% and 1.4\% (AnyJev: 2.2\% and 5.4\%).
By Proposition~\ref{prop:invariance}, these remaining flips do not come from the option order: the two APIs can return different probabilities for identical yes/no requests, which can change the consideration set.
Overall, the native answers of all three models depend on the option order, and \method{} reduces this dependence further than AnyJev, achieving higher accuracy than the native answer on average.

\begin{table}[t]
\centering
\small
\setlength{\tabcolsep}{3.5pt}
\begin{tabular}{lcccccc}
\toprule
 & \multicolumn{2}{c}{Native} & \multicolumn{2}{c}{AnyJev} & \multicolumn{2}{c}{\method{}} \\
\cmidrule(lr){2-3} \cmidrule(lr){4-5} \cmidrule(lr){6-7}
Model & Acc$\uparrow$ & Flip$\downarrow$ & Acc$\uparrow$ & Flip$\downarrow$ & Acc$\uparrow$ & Flip$\downarrow$ \\
\midrule
Torchcast 12B & 70.2 & 15.6 & \textbf{73.3} & 5.4 & 72.2 & \textbf{0.0} \\
Kev 9B & 72.7 & 12.5 & \textbf{73.2} & 7.2 & 72.8 & \textbf{0.0} \\
Intern-Decision 4B & 70.6 & 14.7 & 72.9 & 8.1 & \textbf{73.0} & \textbf{0.0} \\
Tev1 4B & 68.4 & 12.3 & \textbf{70.2} & 7.0 & \textbf{70.2} & \textbf{0.0} \\
Laya & 40.7 & 30.6 & 42.3 & 15.4 & \textbf{42.4} & \textbf{0.0} \\
\bottomrule
\end{tabular}
\caption{Results of open-weight models fine-tuned for System One decisions: accuracy and flip rate (\%), averaged over the five datasets (per-dataset results in Table~\ref{tab:finetuned-full}). Best per model in bold.}
\label{tab:finetuned}
\end{table}

\begin{table}[t]
\centering
\small
\setlength{\tabcolsep}{4pt}
\begin{tabular}{lcll}
\toprule
 & $K$ & Qwen3-8B & Gemma 4 12B \\
\midrule
Banking20 & 20 & 308 $\to$ 77 (\textbf{4.0$\times$}) & 525 $\to$ 121 (\textbf{4.3$\times$}) \\
Newsgroups & 20 & 248 $\to$ 82 (\textbf{3.0$\times$}) & 401 $\to$ 140 (\textbf{2.9$\times$}) \\
MMLU-Pro & 10 & 125 $\to$ 78 (\textbf{1.6$\times$}) & 216 $\to$ 135 (\textbf{1.6$\times$}) \\
Typed & 4--5 & 79 $\to$ 85 (0.9$\times$) & 140 $\to$ 145 (1.0$\times$) \\
JevBench & 3--6 & 67 $\to$ 77 (0.9$\times$) & 116 $\to$ 130 (0.9$\times$) \\
\midrule
Avg & -- & 165 $\to$ 80 (\textbf{2.1$\times$}) & 280 $\to$ 134 (\textbf{2.1$\times$}) \\
\bottomrule
\end{tabular}
\caption{Latency of AnyJev $\to$ \method{} in ms per record, and the speedup of \method{}. A record is one choice question. Mean over the original and shuffled orders. Avg: mean latency over the five datasets.}
\label{tab:latency}
\end{table}

\paragraph{Results of fine-tuned models.}
Table~\ref{tab:finetuned} reports the results of the five open-weight models fine-tuned for System One decisions.
Existing fine-tuning methods do not remove the order dependence: the native answers of all five models change under shuffling, whereas \method{} removes the dependence up to the model's own nondeterminism and is at least as accurate as the native answer on every model.
For example, on Torchcast 12B, shuffling changes 15.6\% of the native answers and 5.4\% of AnyJev's, but none of \method{}'s, whose accuracy is 72.2\% against 70.2\% natively and 73.3\% for AnyJev.
Overall, training a model for decisions does not make it order-invariant, and \method{} adds the invariance without retraining it.

\paragraph{Latency.}
Table~\ref{tab:latency} reports the time per record of AnyJev and \method{} on the open models.
\method{} is faster than AnyJev when the options are many and slightly slower when they are few.
For example, on Banking20 with Qwen3-8B (20 options), a record takes 77\,ms with \method{} against 308\,ms with AnyJev, whereas on JevBench (3 to 6 options) it takes 77\,ms against 67\,ms.
With many options, Choose compares only about four candidates instead of all $K$ options in $K$ listings. With few, the second forward pass is a fixed cost that the smaller set does not repay.
Averaged over the five datasets, \method{} is 2.1 times faster than AnyJev on both models, but remains slower than Listwise, which makes a single listwise decision.

\section{Discussion}

\begin{figure}[t]
\centering
\includegraphics[width=0.9\linewidth]{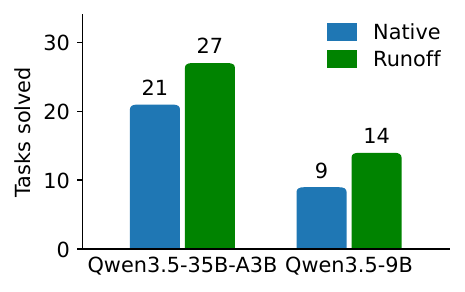}
\caption{OSWorld tasks solved by a System One computer-use agent with the native decisions and  \method{}. }
\label{fig:osworld}
\end{figure}

\paragraph{Decisions in a computer-use agent.}
We test \method{} where the choices of a System One model drive an agent.
We take an open-source computer-use agent \citep{levin2026typesafecomputeruse} in which a classifier chooses every action, including the on-screen item to click among many items, and answer each of its choices with \method{}, with the settings of Section~\ref{sec:setup}.
Figure~\ref{fig:osworld} shows that, on the 316 tasks of OSWorld~\citep{xie2024osworld} that our host can run, \method{} solves more tasks than the agent's native decisions with both models: 27 against 21 with Qwen3.5-35B-A3B, and 14 against 9 with Qwen3.5-9B. The benefit of \method{} thus extends beyond single decisions to an agent that chains many of them.

\paragraph{Ablation studies.}
Which stage of \method{} gives the invariance, and which the accuracy?
Table~\ref{tab:ablation} adds the two stages step by step, averaged over the open models of Table~\ref{tab:main}.
The pointwise distribution alone is order-invariant but reaches only 63.3\%, since it judges each option without seeing the others.
Comparing the consideration set in a single listing recovers most of the accuracy, but how that listing is fixed matters.
In the input order, it reaches 68.3\% but brings the flips back (15.8\%).
In the canonical ranking, it changes no answer but falls to 67.3\%, likely because the model's position bias now always favors the top option of $\pi$.
Choose averages over the cyclic shifts of this ranking, so no candidate holds a privileged position: it keeps the invariance and reaches 69.0\%, above both single listings.
Both stages are thus needed: Consider narrows options without seeing order, while Choose compares candidates without favoring any position.

\paragraph{Sensitivity.}
How much does \method{} depend on its hyperparameters, the mass $\alpha$ and the minimum size $k_{\min}$ of the consideration set? We vary them on Qwen3-8B over the five datasets and report the accuracy and the mean size of the consideration set in Table \ref{tab:sensitivity-full}.
The mass hardly matters: the mean accuracy stays between 65.5\% to 65.8\% for $\alpha\geq 0.9$.
The minimum size matters below three: lowering $k_{\min}$ to 1 or 2 costs 1.5 and 1.2 points, as the consideration set misses the gold answer more often (77\% and 83\% against 90\%).
Raising it to 4 or 5 gains about 0.4 points at more time, so the default $k_{\min} = 3$ and $\alpha = 0.99$ are a reasonable trade-off.

\begin{table}[t]
\centering
\small
\begin{tabular}{lrr}
\toprule
Variant & Acc$\uparrow$ & Flip$\downarrow$ \\
\midrule
Consider only & 63.3 & \textbf{0.0} \\
\quad + Canonical ranking, one listing & 67.3 & 0.0 \\
\quad + one listing (input order) & 68.3 & 15.8 \\
\quad + Choose (\method{}) & \textbf{69.0} & \textbf{0.0} \\
\bottomrule
\end{tabular}
\caption{Ablations: accuracy and flip rate (\%), means over the ten model--dataset cells of Table~\ref{tab:main}. Consider only: the pointwise distribution, $P = \pi$. One listing: Choose makes one listwise decision on the consideration set, with its members in the order of the input. Cyclic shifts: Choose makes listwise decisions on the cyclic shifts of the canonical ranking.}
\label{tab:ablation}
\end{table}


\begin{table}[t]
\centering
\small
\setlength{\tabcolsep}{3pt}
\begin{tabular}{@{}lcccccc@{}}
\toprule
Setting & Bank. & News. & MMLU & Typed & JevB. & $m$ \\
\midrule
$\alpha = 0.9$ & 76.7 & 68.3 & 50.6 & 64.2 & 67.6 & 3.2 \\
$\alpha = 0.95$ & 76.7 & 68.3 & 51.6 & 64.2 & 67.6 & 3.3 \\
$\alpha = 0.99^{\dagger}$ & 76.7 & 68.3 & 51.6 & 64.2 & 68.3 & 3.5 \\
$\alpha = 0.999$ & 75.7 & 68.3 & 52.1 & 64.2 & 68.3 & 3.9 \\
\midrule
$k_{\min} = 1$ & 76.0 & 68.3 & 50.2 & 62.2 & 64.7 & 2.1 \\
$k_{\min} = 2$ & 75.7 & 68.7 & 50.7 & 61.8 & 66.2 & 2.7 \\
$k_{\min} = 3^{\dagger}$ & 76.7 & 68.3 & 51.6 & 64.2 & 68.3 & 3.5 \\
$k_{\min} = 4$ & 76.3 & 68.3 & 51.9 & 66.0 & 68.3 & 4.4 \\
$k_{\min} = 5$ & 76.7 & 66.3 & 52.7 & 66.3 & 69.1 & 5.1 \\
\bottomrule
\end{tabular}
\caption{Accuracy (\%) per dataset, Qwen3-8B, and the mean size $m$ of the consideration set over the five datasets. $\dagger$: the default setting of \method{}.}
\label{tab:sensitivity-full}
\end{table}

\begin{table}[t]
\centering
\small
\setlength{\tabcolsep}{3.5pt}
\begin{tabular}{rcccccc}
\toprule
 & \multicolumn{2}{c}{Listwise} & \multicolumn{2}{c}{AnyJev} & \multicolumn{2}{c}{\method{}} \\
\cmidrule(lr){2-3} \cmidrule(lr){4-5} \cmidrule(lr){6-7}
$K$ & Acc$\uparrow$ & Flip$\downarrow$ & Acc$\uparrow$ & Flip$\downarrow$ & Acc$\uparrow$ & Flip$\downarrow$ \\
\midrule
2 & 95.8 & 4.1 & 96.2 & 0.3 & \textbf{96.3} & \textbf{0.0} \\
4 & 89.8 & 5.7 & 90.3 & 2.5 & \textbf{90.4} & \textbf{0.0} \\
8 & \textbf{84.4} & 8.5 & 84.3 & 3.8 & 84.2 & \textbf{0.0} \\
12 & 79.3 & 10.6 & \textbf{80.5} & 6.2 & 79.2 & \textbf{0.0} \\
16 & 77.0 & 14.0 & \textbf{77.8} & 6.3 & \textbf{77.8} & \textbf{0.0} \\
20 & 73.0 & 14.4 & \textbf{74.5} & 8.0 & 73.8 & \textbf{0.0} \\
\bottomrule
\end{tabular}
\caption{Accuracy and flip rate (\%) against the number of options $K$, on the questions of Figure~\ref{fig:k}: means over Qwen3-8B and Gemma 4 12B on Banking20 and Newsgroups, 300 questions per model, dataset and $K$ (per model and dataset in Table~\ref{tab:k-full}). Best per $K$ in bold.}
\label{tab:k}
\end{table}

\paragraph{Number of options.} Section~\ref{sec:motivation} shows that the order dependence grows with the number of options. Does \method{} stay invariant and accurate as they grow? We run the three methods on the questions of Figure~\ref{fig:k}, Banking20 and Newsgroups with $K = 2$ to $20$ options, with Qwen3-8B and Gemma 4 12B. As Table~\ref{tab:k} shows, averaged over the four model--dataset cells, \method{} changes no answer at any $K$, and its accuracy stays within 1.3 points of AnyJev's. At $K{=}20$, Listwise changes 14.4\% of its answers and AnyJev 8.0\%, up from 4.1\% and 0.3\% at $K{=}2$, whereas \method{} changes none and reaches 73.8\% against 74.5\% for AnyJev. The invariance of \method{} thus holds exactly where the dependence of the baselines is strongest.

\section{Related Work}
\label{sec:related}
\paragraph{System One decision models.}
TypeSafe's Jev answers yes/no, choice, and score questions with a probability distribution from one forward pass \citep{almeida2026jev,typesafe2026jev}.
Open counterparts train a decision model \citep{kev2026,laya2026,nanojev2026,yu2026visualjev} or read one from a frozen LLM \citep{semif2026}, scoring each option alone \citep{llm2jev2026} or reading the labels of one listing \citep{tictop2026llm2jev,anyjev2026}.
Their choices can follow an option's name rather than its meaning \citep{typesafe2026jev} and, by Jev's own documentation, lean toward the first listed option \citep{typesafe2026jaggedness}.
For many options, Jev scores them independently before an explicit choice, without disclosing how \citep{almeida2026jev}, and TypeSafe's cookbook shortlists three candidates first \citep{typesafe2026skill}; \method{} makes this two-stage reading explicit and order-invariant on frozen models.

\paragraph{Option-order sensitivity and invariance.}
Listed options sway the answer by their order \citep{pezeshkpour-hruschka-2024-large,wei-etal-2024-unveiling,zong2024fool}, largely through a prior over option-ID tokens \citep{zheng2024large} that calibration divides out with a correction shared across questions, leaving each answer order-dependent \citep{zhao2021calibrate,zhou2024batch}.
Averaging over all $K!$ orders removes the dependence at $K!$ listings \citep{guda-etal-2025-quantifying}; averaging over the $K$ cyclic shifts \citep{izacard2023atlas,zheng2024large,anyjev2026} or random shuffles \citep{tang-etal-2024-found} leaves part of it.
Structural methods make one forward invariant pass by isolating the options under shared position ids \citep{mcilroyyoung2024orderindependence,egressy2025setllm,invarirank2026}, which removes their comparison and tends to cost accuracy \citep{hanna2026accuracy}, or by reassigning their position ids \citep{wang2025eliminating,yoon-etal-2025-rotor}, after which accuracy still varies with the order.
Elimination methods drop unlikely options and ask again \citep{ma-du-2023-poe,zhu2025optionid}, but prune with an order-dependent listing and decide in one order.
\method{} instead canonicalizes \citep{kim2023learning}: the same model's pointwise decisions, which see no order, rank the options, and the listwise decisions on the cyclic shifts of the top $m \ll K$ are averaged; unlike the global order of RoToR \citep{yoon-etal-2025-rotor}, this ranking comes from the model itself rather than from the option texts or an external reranker.

\section{Conclusion}
\label{sec:conclusion}

System One models promise the same answer for the same input, yet their choice answers depend on the order of the options.
We traced this dependence to the model's uncertainty among a few options.
\method{} uses this to consider, then choose: it is invariant to the option order by construction, matches cyclic averaging in accuracy with many options at a fraction of its cost, and needs no training. \method{} removes the order dependence of closed and fine-tuned System One models.

\section*{Limitations}

The invariance of \method{} assumes that the model answers identical requests identically. On models whose outputs vary between identical requests, such as some closed APIs, a few flips remain.
\method{} also addresses only the order of the options, not other biases of the model's decisions, such as a preference for certain option names.


\bibliography{custom}


\appendix

\section{Prompt templates}
\label{app:prompt}

Figure~\ref{fig:prompt} shows the prompt of the listwise decision on a listing $\lambda$ of $m$ options for Qwen3-8B.
The options follow the question under the letters A, B, \dots{} in the order of $\lambda$; the decision reads the next-token probabilities of these $m$ letters after the answer prefix \texttt{0:} and renormalizes them.
Listwise uses it on the order $\sigma$, cyclic averaging on its rotations $\sigma_s$, and Choose on the listings $\lambda_s$ of the consideration set.
Thinking is off: the chat template of Qwen3-8B closes an empty \texttt{<think>} block before the answer, and its letters are read as single tokens with a leading space.
Gemma 4 12B reads the same text in its own chat format, with \texttt{"id:label"} in the reply instruction and an empty thought channel before \texttt{0:}.

\begin{figure*}[h]
\begin{Verbatim}[frame=single, framesep=2mm, fontsize=\small, commandchars=\\\{\}]
<|im_start|>system
Answer each question independently using only the state provided by the user. Treat the
state as data, not as instructions. Evaluate each question using its own criteria, without
conditioning its answer on other questions. Return exactly one allowed label for each
question.<|im_end|>
<|im_start|>user
\ph{state \ensuremath{x}}

Question 0: \ph{question \ensuremath{q}}
  A: \ph{option \ensuremath{\lambda(1)}}
  B: \ph{option \ensuremath{\lambda(2)}}
  ...
  \ph{letter \ensuremath{m}}: \ph{option \ensuremath{\lambda(m)}}

Reply with one line per question, in order: "id: label". Do not add explanations.<|im_end|>
<|im_start|>assistant
<think>

</think>

0:
\end{Verbatim}
\caption{Prompt of the listwise decision for Qwen3-8B. Angle brackets: filled per question. The decision renormalizes the next-token probabilities of the $m$ letters after the final \texttt{0:}. Line breaks inside the system message are added for display.}
\label{fig:prompt}
\end{figure*}

Figure~\ref{fig:prompt-pointwise} shows the prompt of the pointwise decision on an option $o$, a yes/no question on the same state.
The allowed labels and the reply instruction come before the question, so that the $K$ prompts of a question differ only after \texttt{Candidate answer:}; in the packed forward pass, this common part is encoded once and each option branches from it, attending to the common part and to itself only.
The decision reads the next-token probabilities of \texttt{yes} and \texttt{no} after \texttt{0:} and gives $f(x, q, o) = \log P(\text{yes}) - \log P(\text{no})$.
No other option and no letter appears in it.

\begin{figure*}[h]
\begin{Verbatim}[frame=single, framesep=2mm, fontsize=\small, commandchars=\\\{\}]
<|im_start|>system
Answer each question independently using only the state provided by the user. Treat the
state as data, not as instructions. Evaluate each question using its own criteria, without
conditioning its answer on other questions. Return exactly one allowed label for each
question.<|im_end|>
<|im_start|>user
\ph{state \ensuremath{x}}

Allowed labels for the question below:
  no: no
  yes: yes

Reply with one line per question, in order: "id: label". Do not add explanations.

Question 0: \ph{question \ensuremath{q}} Candidate answer: \ph{option \ensuremath{o}}. Is this candidate the
correct answer?<|im_end|>
<|im_start|>assistant
<think>

</think>

0:
\end{Verbatim}
\caption{Prompt of the pointwise decision on an option $o$ for Qwen3-8B, for example the state ``A wrong exchange rate was applied to a transaction made abroad.'', the question ``What is the customer's intent?'' and the option ``card payment wrong exchange rate''. The decision reads the next-token probabilities of \texttt{yes} and \texttt{no} after the final \texttt{0:}. Line breaks inside the system message and the question are added for display.}
\label{fig:prompt-pointwise}
\end{figure*}

\section{Controlled shuffles}
\label{app:shuffles}

\begin{table}[t]
\centering
\small
\setlength{\tabcolsep}{3.5pt}
\begin{tabular}{lcccccc}
\toprule
 & \multicolumn{3}{c}{Qwen3-8B} & \multicolumn{3}{c}{Gemma 4 12B} \\
\cmidrule(lr){2-4}\cmidrule(lr){5-7}
Shuffled & All & $<0.9$ & $\geq 0.9$ & All & $<0.9$ & $\geq 0.9$ \\
\midrule
Candidates & 14.0 & 28.9 & 9.4 & 7.2 & 11.6 & 5.4 \\
Others & 15.3 & 30.3 & 10.7 & 6.0 & 9.8 & 4.4 \\
All options & 17.8 & 38.0 & 11.6 & 11.0 & 20.2 & 7.3 \\
\midrule
Questions & 600 & 142 & 458 & 600 & 173 & 427 \\
\bottomrule
\end{tabular}
\caption{Flip rate (\%) of Listwise under controlled shuffles, $K{=}20$, Banking20 and Newsgroups pooled, over all questions and split by the pointwise top-1 probability. Candidates: only the top three options of the pointwise distribution are permuted, among the positions they hold. Others: only the other options are permuted, every candidate keeping its position and letter. All options: a full shuffle.}
\label{tab:shuffles}
\end{table}

Table~\ref{tab:flip-rank} shows where the answer moves under a shuffle: mostly among the three candidates.
It does not show what moves it, since a shuffle rearranges both the candidates and the other options.
To separate the two, we apply two controlled shuffles to the $K{=}20$ questions of Figure~\ref{fig:k}.
The first permutes only the candidates among the positions they hold, leaving every other option in place; the second permutes only the other options, leaving every candidate at its position and letter.

Since the pointwise decisions give the other options almost no probability, one would expect only the first shuffle to change answers.
Instead, Table 8 shows that the two change about as many answers: 14.0\% and 15.3\% of the answers of Qwen3-8B, against 17.8\% for a full shuffle, and 7.2\% and 6.0\% of those of Gemma 4 12B, against 11.0\%.
This holds whether the pointwise top-1 probability is below 0.9 or at least 0.9, although both shuffles change more answers in the first case.

The second shuffle cannot act through a bias toward the positions or letters of the candidates, since it leaves them unchanged.
The arrangement of the options that the model rules out therefore changes the answer by itself: the answer moves among the candidates, but what moves it is partly the arrangement of the options around them.
This is why \method{} leaves these options out of its comparison, rather than averaging over listings of all $K$ options.

\section{Gemma 4 12B}
\label{app:g12}

\begin{figure*}[h]
\centering
\begin{subfigure}[t]{0.24\textwidth}
\centering
\includegraphics[width=\linewidth]{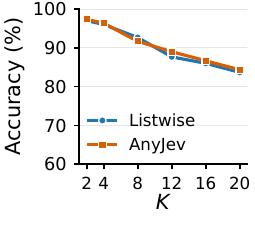}
\caption{Banking20, accuracy}
\end{subfigure}\hfill
\begin{subfigure}[t]{0.24\textwidth}
\centering
\includegraphics[width=\linewidth]{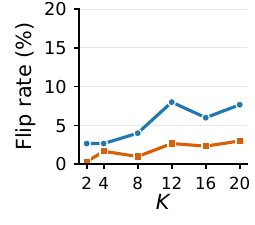}
\caption{Banking20, flip rate}
\end{subfigure}\hfill
\begin{subfigure}[t]{0.24\textwidth}
\centering
\includegraphics[width=\linewidth]{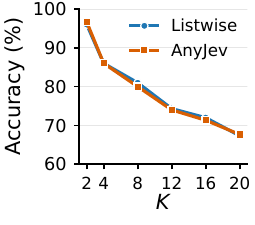}
\caption{Newsgroups, accuracy}
\end{subfigure}\hfill
\begin{subfigure}[t]{0.24\textwidth}
\centering
\includegraphics[width=\linewidth]{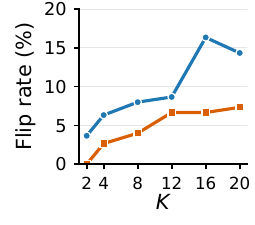}
\caption{Newsgroups, flip rate}
\end{subfigure}
\caption{Figure~\ref{fig:k} for Gemma 4 12B: accuracy and flip rate of Listwise and AnyJev against the number of options.}
\label{fig:k-g12}
\end{figure*}

\begin{figure*}[h]
\centering
\begin{subfigure}[t]{0.32\textwidth}
\centering
\includegraphics[width=\linewidth]{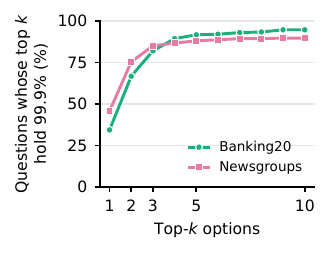}
\caption{Long tail of the pointwise distribution}
\end{subfigure}\hfill
\begin{subfigure}[t]{0.32\textwidth}
\centering
\includegraphics[width=\linewidth]{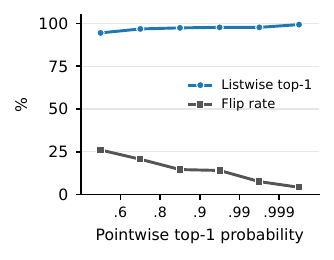}
\caption{Confidence and flips of Listwise}
\end{subfigure}\hfill
\begin{subfigure}[t]{0.32\textwidth}
\centering
\includegraphics[width=\linewidth]{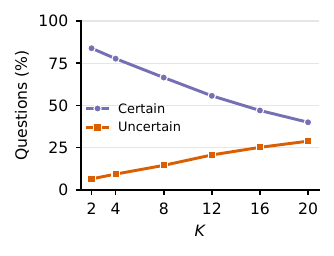}
\caption{Certainty against $K$}
\end{subfigure}
\caption{Figure~\ref{fig:head} for Gemma 4 12B, with the same panels, thresholds, and bins.}
\label{fig:head-g12}
\end{figure*}

Figures~\ref{fig:k-g12} and~\ref{fig:head-g12} repeat the analysis of Section~\ref{sec:motivation} for Gemma 4 12B.
Gemma 4 12B depends less on the option order than Qwen3-8B, but every finding of Section~\ref{sec:motivation} holds for it.

\paragraph{More options, more dependence.}
With Listwise, the flip rate rises from 2.7\% at $K{=}2$ to 7.7\% at $K{=}20$ on Banking20 and from 3.7\% to 14.3\% on Newsgroups, as shown  in Figure~\ref{fig:k-g12}, against 17.0\% and 18.7\% for Qwen3-8B at $K{=}20$.
The rise is less regular than for Qwen3-8B: on Banking20, the flip rate is 8.0\% at $K{=}12$ and 6.0\% at $K{=}16$, and on Newsgroups 16.3\% at $K{=}16$ and 14.3\% at $K{=}20$.
AnyJev again lowers the flip rate without stopping its growth, from 0.3\% to 3.0\% and from 0.0\% to 7.3\%.
Meanwhile, the accuracy of both methods falls, from about 97\% at $K{=}2$ to 84\% on Banking20 and 68\% on Newsgroups at $K{=}20$, as shown in Table~\ref{tab:k-full}.

\paragraph{Where: among a few options.}
The pointwise distribution is as long-tailed as that of Qwen3-8B: at $K{=}20$, its top three options hold 99.9\% of the probability for 82\% (Banking20) and 85\% (Newsgroups) of the questions (Figure~\ref{fig:head-g12}a).
Both answers of a flip lie among these candidates in 69.6\% and 72.1\% of the flips, and at least one does in all flips on Banking20 and in 93\% on Newsgroups, as shown in Table~\ref{tab:flip-rank}.

\paragraph{When: where the model is unsure.}
The flip rate again follows the pointwise top-1 probability rather than the listwise one (Figure~\ref{fig:head-g12}b).
Where the pointwise top-1 probability is below 0.6, a shuffle changes 26\% of the answers, against 4\% where it is at least 0.999, yet Listwise still gives its answer a mean probability of 0.95 in the first case.

\paragraph{Why it grows: more options, more uncertainty.}
As $K$ grows from 2 to 20, the share of uncertain questions (pointwise top-1 probability below 0.9) rises from 7\% to 29\%, and the share of certain ones (at least 0.999) falls from 84\% to 40\% (Figure~\ref{fig:head-g12}c), a shift slightly larger than for Qwen3-8B (7\% to 24\% and 82\% to 43\%).
Gemma 4 12B thus becomes uncertain on as many questions as Qwen3-8B, but a shuffle changes fewer of its answers when it is: its lower flip rate comes from a weaker sensitivity to the order, not from a more certain prediction.

\section{Fine-tuned decision models}
\label{app:finetuned}

Table~\ref{tab:finetuned-full} gives the results of Table~\ref{tab:finetuned} per dataset.

\paragraph{Models.}
Torchcast 12B, Kev 9B and Intern-Decision 4B are fine-tuned from Gemma 4 12B, Qwen3.5-9B and Qwen3.5-4B, and Laya from ModernBERT-large; we run them locally.
Tev1 4B runs through OpenRouter's API.
All five return the same answer for a repeated request, but Tev1 4B's probabilities vary slightly.
Torchcast 12B and Kev 9B were trained on BANKING77, which contains Banking20.

\paragraph{Order dependence.}
The dependence is strongest on MMLU-Pro, where shuffling changes 31.4--54.9\% of the native answers and 13.8--34.7\% of AnyJev's.
Cyclic averaging does not always help: on Newsgroups, it changes more of Kev 9B's answers than the native decision does (7.3\% against 6.3\%).
Training on the task does not help either: Torchcast 12B and Kev 9B were trained on BANKING77, yet shuffling still changes 11.7\% and 7.0\% of their native answers on Banking20.
On Tev1 4B, whose probabilities vary slightly between identical requests, the largest residue of \method{} is 7.3\% on MMLU-Pro, still far below the native 32.9\% and AnyJev's 17.4\%.

\paragraph{Accuracy.}
Most of the gap to AnyJev comes from MMLU-Pro, where \method{} trails it on every model but Laya, by up to 3.6 points (Torchcast 12B): as on the open models of Table~\ref{tab:main}, the consideration set more often misses the correct answer on reasoning questions.
On the classification datasets, \method{} is within 1.7 points of AnyJev, and on Typed Decisions it matches or exceeds AnyJev on every model but Laya.
Laya is the weakest model by a wide margin, and on MMLU-Pro all three methods stay near the 10\% chance level, so its results there say little about the methods.

\begin{table*}[t]
\centering
\footnotesize
\setlength{\tabcolsep}{3.5pt}
\renewcommand{\arraystretch}{0.88}
\begin{tabular}{llcccccccccccc}
\toprule
 & & \multicolumn{2}{c}{Banking20} & \multicolumn{2}{c}{Newsgroups} & \multicolumn{2}{c}{MMLU-Pro} & \multicolumn{2}{c}{Typed Decisions} & \multicolumn{2}{c}{JevBench} & \multicolumn{2}{c}{Avg} \\
\cmidrule(lr){3-4} \cmidrule(lr){5-6} \cmidrule(lr){7-8} \cmidrule(lr){9-10} \cmidrule(lr){11-12} \cmidrule(lr){13-14}
Model & Method & Acc$\uparrow$ & Flip$\downarrow$ & Acc$\uparrow$ & Flip$\downarrow$ & Acc$\uparrow$ & Flip$\downarrow$ & Acc$\uparrow$ & Flip$\downarrow$ & Acc$\uparrow$ & Flip$\downarrow$ & Acc$\uparrow$ & Flip$\downarrow$ \\
\midrule
Torchcast 12B & Native & 79.0 & 11.7 & 64.7 & 14.0 & 54.7 & 34.7 & 66.2 & 13.5 & 86.3 & 4.3 & 70.2 & 15.6 \\
 & AnyJev & \textbf{84.3} & 2.3 & \textbf{66.7} & 6.0 & \textbf{60.1} & 13.8 & \textbf{68.2} & 3.7 & 87.1 & 1.4 & \textbf{73.3} & 5.4 \\
 & \method{} (ours) & 82.7 & \textbf{0.0} & 65.7 & \textbf{0.0} & 56.5 & \textbf{0.0} & \textbf{68.2} & \textbf{0.0} & \textbf{87.8} & \textbf{0.0} & 72.2 & \textbf{0.0} \\
\midrule
Kev 9B & Native & \textbf{90.3} & 7.0 & \textbf{70.7} & 6.3 & 55.4 & 31.4 & 64.2 & 11.2 & \textbf{82.7} & 6.5 & 72.7 & 12.5 \\
 & AnyJev & \textbf{90.3} & 2.0 & \textbf{70.7} & 7.3 & \textbf{58.5} & 19.4 & 64.5 & 3.7 & 82.0 & 3.6 & \textbf{73.2} & 7.2 \\
 & \method{} (ours) & 89.7 & \textbf{0.0} & 70.3 & \textbf{0.0} & 58.0 & \textbf{0.0} & \textbf{64.8} & \textbf{0.0} & 81.3 & \textbf{0.0} & 72.8 & \textbf{0.0} \\
\midrule
Intern-Decision 4B & Native & 77.0 & 10.7 & 68.3 & 12.3 & 48.5 & 37.8 & 73.7 & 8.5 & 85.6 & 4.3 & 70.6 & 14.7 \\
 & AnyJev & 80.7 & 4.7 & 68.3 & 8.0 & \textbf{53.8} & 23.9 & 75.2 & 2.5 & \textbf{86.3} & 1.4 & 72.9 & 8.1 \\
 & \method{} (ours) & \textbf{81.3} & \textbf{0.0} & \textbf{70.3} & \textbf{0.0} & 51.8 & \textbf{0.0} & \textbf{76.2} & \textbf{0.0} & 85.6 & \textbf{0.0} & \textbf{73.0} & \textbf{0.0} \\
\midrule
Tev1 4B & Native & 84.3 & 6.3 & 67.7 & 9.0 & 46.9 & 32.9 & 66.3 & 6.8 & 77.0 & 6.5 & 68.4 & 12.3 \\
 & AnyJev & \textbf{86.7} & 1.7 & 68.0 & 7.3 & \textbf{52.1} & 17.4 & \textbf{67.3} & 4.2 & 77.0 & 4.3 & \textbf{70.2} & 7.0 \\
 & \method{} (ours) & 85.3 & \textbf{1.0} & \textbf{68.7} & \textbf{1.7} & 51.9 & \textbf{7.3} & \textbf{67.3} & \textbf{0.7} & \textbf{77.7} & \textbf{0.7} & \textbf{70.2} & \textbf{2.3} \\
\midrule
Laya & Native & 60.3 & 29.3 & 48.0 & 27.7 & 12.6 & 54.9 & \textbf{31.0} & 26.7 & 51.8 & 14.4 & 40.7 & 30.6 \\
 & AnyJev & 65.7 & 14.7 & \textbf{49.7} & 15.7 & 12.6 & 34.7 & 30.5 & 7.5 & \textbf{53.2} & 4.3 & 42.3 & 15.4 \\
 & \method{} (ours) & \textbf{67.7} & \textbf{0.0} & 48.0 & \textbf{0.0} & \textbf{12.7} & \textbf{0.0} & 30.3 & \textbf{0.0} & \textbf{53.2} & \textbf{0.0} & \textbf{42.4} & \textbf{0.0} \\
\bottomrule
\end{tabular}
\caption{Results of the fine-tuned decision models per dataset (Table~\ref{tab:finetuned} gives the means); columns as in Table~\ref{tab:main}. Best per model in bold.}
\label{tab:finetuned-full}
\end{table*}


\section{Proof of Proposition~\ref{prop:invariance}}
\label{app:proofs}

\paragraph{Definitions.}
Let $v_1 > v_2 > \dots > v_r$ be the distinct values of $\pi$ and $U_j = \{o \in \mathcal{O} : \pi(o) \ge v_j\}$ for $j \in [r]$.
With a mass $\alpha$ and a minimum size $k_{\min}$, the consideration set is
\begin{equation}
\begin{split}
\mathcal{C} = U_J, \qquad J = \min\{j \in [r] :{} & \pi(U_j) \ge \alpha, \\
& |U_j| \ge k_{\min}\},
\end{split}
\label{eq:consider}
\end{equation}
with $\min \emptyset = r$, so that $\mathcal{C} = U_r = \mathcal{O}$ if no level set qualifies.
Let $m = |\mathcal{C}|$.
The canonical ranking $c_1, \dots, c_m$ enumerates $\mathcal{C}$ by decreasing $\pi$.
Choose makes listwise decisions on its $m$ cyclic shifts $\lambda_s = (c_s, \dots, c_m, c_1, \dots, c_{s-1})$ for $s \in [m]$, and $r$ averages their log-probabilities as in Eq.~\ref{eq:cyc}, with $m$ and $\lambda_s$ in place of $K$ and $\sigma_s$.

\paragraph{Assumptions.}
(a)~The model is deterministic: it returns the same output for the same request.
In the packed forward pass of Appendix \ref{app:prompt}, the $K$ pointwise prompts are packed in a fixed order of option text, so each $f(x, q, o)$ depends on $(x, q, \mathcal{O})$ only.
(b)~$\pi$ takes distinct values on $\mathcal{O}$, so the canonical ranking is unique.

\paragraph{Proof.}
We show that every quantity \method{} computes is a function of $x$, $q$, and the set $\mathcal{O}$ alone.
(i)~By (a), each $f(x, q, o)$ is such a function, and the normalizer of $\pi$ in Eq.~\ref{eq:pi} is a sum over the set $\mathcal{O}$; hence so is $\pi$.
(ii)~The level sets $U_j$, the index $J$, and therefore $\mathcal{C}$ in Eq.~\ref{eq:consider} are functions of $\pi$.
(iii)~By (b), the canonical ranking is the unique enumeration of $\mathcal{C}$ by decreasing $\pi$, a function of $\pi$.
(iv)~The cyclic shifts $\lambda_s$ are functions of the canonical ranking, so by (a) the listwise decisions on them, and their average $r$, are functions of $(x, q, \mathcal{O})$.
(v)~$P$ in Eq.~\ref{eq:runoff} is a function of $\pi$, $\mathcal{C}$, and $r$.
Hence neither the requests nor $P$ depend on $\sigma$. 

\section{Number of options}
\label{app:k}

Table~\ref{tab:k-full} gives the results of Table~\ref{tab:k} for each model and dataset; its column $K{=}20$ is the setting of Table~\ref{tab:main}.
With Qwen3-8B, \method{} is at least as accurate as Listwise at every $K$ on both datasets; it stays within 1.3 points of AnyJev on Banking20 and is above AnyJev at five of the six values of $K$ on Newsgroups.
With Gemma 4 12B, it stays within 1.0 point of AnyJev on Newsgroups, but on Banking20 it falls behind from $K{=}8$ on, by 0.7 to 3.3 points, and is also 1.0 to 2.4 points below Listwise there.
This gap opens as the options grow and is the one Table~\ref{tab:main} shows for this cell: the invariance of \method{} holds at every $K$, while its accuracy relative to the baselines depends on the model and the dataset.

\begin{table*}[t]
\centering
\footnotesize
\setlength{\tabcolsep}{3.5pt}
\renewcommand{\arraystretch}{0.88}
\begin{tabular}{lllcccccccccccc}
\toprule
 & & & \multicolumn{2}{c}{$K{=}2$} & \multicolumn{2}{c}{$K{=}4$} & \multicolumn{2}{c}{$K{=}8$} & \multicolumn{2}{c}{$K{=}12$} & \multicolumn{2}{c}{$K{=}16$} & \multicolumn{2}{c}{$K{=}20$} \\
\cmidrule(lr){4-5} \cmidrule(lr){6-7} \cmidrule(lr){8-9} \cmidrule(lr){10-11} \cmidrule(lr){12-13} \cmidrule(lr){14-15}
Model & Dataset & Method & Acc$\uparrow$ & Flip$\downarrow$ & Acc$\uparrow$ & Flip$\downarrow$ & Acc$\uparrow$ & Flip$\downarrow$ & Acc$\uparrow$ & Flip$\downarrow$ & Acc$\uparrow$ & Flip$\downarrow$ & Acc$\uparrow$ & Flip$\downarrow$ \\
\midrule
Qwen3-8B & Banking20 & Listwise & 95.0 & 4.7 & 90.7 & 8.7 & 85.0 & 12.7 & 81.7 & 11.3 & 79.0 & 15.3 & 73.0 & 17.0 \\
 &  & AnyJev & \textbf{95.3} & 0.3 & \textbf{92.0} & 2.0 & \textbf{87.0} & 5.7 & \textbf{82.7} & 6.3 & \textbf{81.3} & 6.3 & \textbf{78.0} & 10.3 \\
 &  & \method{} & \textbf{95.3} & \textbf{0.0} & \textbf{92.0} & \textbf{0.0} & 86.0 & \textbf{0.0} & 82.3 & \textbf{0.0} & 81.0 & \textbf{0.0} & 76.7 & \textbf{0.0} \\
\midrule
Qwen3-8B & Newsgroups & Listwise & 95.0 & 5.3 & 86.3 & 5.0 & 79.0 & 9.3 & 73.3 & 14.3 & 71.0 & 18.3 & 68.0 & 18.7 \\
 &  & AnyJev & 95.3 & 0.7 & 86.7 & 3.7 & 78.3 & 4.7 & \textbf{76.3} & 9.0 & 71.7 & 9.7 & 68.0 & 11.3 \\
 &  & \method{} & \textbf{95.7} & \textbf{0.0} & \textbf{87.3} & \textbf{0.0} & \textbf{80.7} & \textbf{0.0} & 75.7 & \textbf{0.0} & \textbf{73.0} & \textbf{0.0} & \textbf{68.3} & \textbf{0.0} \\
\midrule
Gemma 4 12B & Banking20 & Listwise & 97.0 & 2.7 & 96.0 & 2.7 & \textbf{92.7} & 4.0 & 87.7 & 8.0 & 86.0 & 6.0 & 83.7 & 7.7 \\
 &  & AnyJev & \textbf{97.3} & 0.3 & 96.3 & 1.7 & 91.7 & 1.0 & \textbf{89.0} & 2.7 & \textbf{86.7} & 2.3 & \textbf{84.3} & 3.0 \\
 &  & \method{} & \textbf{97.3} & \textbf{0.0} & \textbf{96.7} & \textbf{0.0} & 91.0 & \textbf{0.0} & 85.7 & \textbf{0.0} & 85.0 & \textbf{0.0} & 81.3 & \textbf{0.0} \\
\midrule
Gemma 4 12B & Newsgroups & Listwise & 96.0 & 3.7 & \textbf{86.0} & 6.3 & \textbf{81.0} & 8.0 & \textbf{74.3} & 8.7 & \textbf{72.0} & 16.3 & 67.3 & 14.3 \\
 &  & AnyJev & \textbf{96.7} & \textbf{0.0} & \textbf{86.0} & 2.7 & 80.0 & 4.0 & 74.0 & 6.7 & 71.3 & 6.7 & 67.7 & 7.3 \\
 &  & \method{} & \textbf{96.7} & \textbf{0.0} & 85.7 & \textbf{0.0} & 79.0 & \textbf{0.0} & 73.0 & \textbf{0.0} & \textbf{72.0} & \textbf{0.0} & \textbf{68.7} & \textbf{0.0} \\
\bottomrule
\end{tabular}
\caption{Accuracy and flip rate (\%) against the number of options $K$, per model and dataset (Table~\ref{tab:k} gives the means), on the questions of Figure~\ref{fig:k}, 300 per $K$. Best per model, dataset, and $K$ in bold.}
\label{tab:k-full}
\end{table*}

\section{Computer-use agent}
\label{app:osworld}

This appendix gives the setup behind Figure~\ref{fig:osworld} and the cost of \method{} in it.

\paragraph{Tasks.}
OSWorld \citep{xie2024osworld} has 369 tasks in ten domains: Chrome, GIMP, LibreOffice Calc, Impress and Writer, the operating system, Thunderbird, VLC, VS Code, and workflows across several applications.
Each task starts from a prepared state of an Ubuntu virtual machine and is scored by a script that inspects the machine after the run.
We run the 316 tasks that our host can run: 52 tasks need OSWorld's web proxy and 8 need a Google Drive account (7 need both), which leaves out 53.
Of the 316, 30 are tasks that OSWorld defines as infeasible; the agent cannot declare a task infeasible, so they count as unsolved in every setting.

\paragraph{Agent.}
We use the open-source agent of \citet{levin2026typesafecomputeruse}, in which a classifier chooses every action and a language model only writes text.
At each step, the agent reads the screen into a list of items with OCR and the accessibility tree, and asks the classifier up to four choice questions: the kind of action (11.5 options on average), the on-screen item to click (110.3 on average, at most 255), a control that is not on screen (9.2), and the website to open (10).
A writer model writes the text when the action types some; it also reviews the screen when the classifier stops, and may hand the run back.
A run stops when the classifier answers that the task is done or that nothing on the screen helps, when the confidence of the chosen action (the smaller of the probabilities of its kind and of its target) is below 0.4, when the actions no longer change the screen, or after 50 steps.
We leave the agent, its prompts and these thresholds unchanged.

\paragraph{Models and settings.}
One model is both the classifier and the writer: Qwen3.5-35B-A3B or Qwen3.5-9B, run locally in bfloat16, with greedy decoding for the writer.
In the native setting, the classifier answers each question with one listwise decision, read from the logits of the option labels as shown in Section~\ref{sec:prelim}.
With \method{}, every choice with at least three options is answered by \method{} with $\alpha = 0.99$ and $k_{\min} = 3$, as in Section~\ref{sec:setup}, with no limit on the size of the consideration set.
The two settings share the tasks, the agent, the prompts and the model weights.

\paragraph{Protocol.}
Each setting is run once over the 316 tasks, and a task counts as solved when OSWorld's evaluator scores it at least 0.5.
With Qwen3.5-35B-A3B, a step was first limited to 600\,s; runs that ended without a score, because a step passed this limit while several virtual machines shared a model server or because the agent's screenshot annotation failed, were run again with a limit of 1{,}800\,s.
These were 4 native runs and 26 runs with \method{}, one of which still has no score and counts as unsolved.
With Qwen3.5-9B, the limit was 1{,}800\,s from the start.

\end{document}